\documentclass{article}
\usepackage{graphicx}
\usepackage{authblk}
\usepackage{url}
\usepackage{amsmath}    % 数学符号支持
\usepackage{booktabs}   % 表格优化
\usepackage{caption}    % 图表标题优化
\usepackage{float}      % 浮动体控制
\usepackage{placeins}   % 用于 \FloatBarrier
\usepackage[T1]{fontenc}
\usepackage[utf8]{inputenc}

\title{Diagnosing Hallucination Risk in AI Surgical Decision-Support: A Sequential Framework for Sequential Validation}

\author[1,2,3]{Dong Chen}
\author[1]{Yanzhe Wei}
\author[3]{Zonglin He}
\author[1]{Guan-Ming Kuang}
\author[1]{Canhua Ye}
\author[3]{Meiru An}
\author[1]{Huili Peng}
\author[1,3]{Yong Hu}
\author[1,3]{Huiren Tao}
\author[1,2,3]{Kenneth MC Cheung\thanks{Corresponding author, Email: cheungmc@hku.hk}}
\affil[1]{Orthopaedic Centre, The University of Hong Kong - Shenzhen Hospital, Shenzhen, China}
\affil[2]{Translational Medicine Centre, The University of Hong Kong - Shenzhen Hospital, China}
\affil[3]{Department of Orthopaedics \& Traumatology, Li Ka Shing Faculty of Medicine, The University of Hong Kong, Hong Kong, China}

\date{October 2025}

\begin{document}

\maketitle

\begin{abstract}
Large language models (LLMs) hold transformative potential in clinical decision-support but introduce significant risks through hallucinations, which are factually inconsistent or contextually misaligned outputs that may compromise patient safety. This study introduces a novel, clinician-centered framework to quantify and mitigate hallucination risks in spine surgery by assessing diagnostic precision, recommendation quality, reasoning robustness, output coherence, and knowledge alignment. Our evaluation of six leading LLMs across 30 expert-validated Chinese spinal cases revealed robust inter-examiner reliability (mean r = 0.90 ± 0.014) among three independent surgeon-evaluators. DeepSeek-R1 demonstrated superior overall performance (total score: 86.03 ± 2.08), particularly in high-stakes domains such as trauma and infection. A critical finding emerged that reasoning-enhanced model variants did not uniformly outperform their standard counterparts. Notably, the extended thinking mode of Claude-3.7-Sonnet underperformed relative to its standard version (80.79 ± 1.83 vs. 81.56 ± 1.92), underscoring that extended chain-of-thought reasoning alone is insufficient for ensuring clinical reliability. Multidimensional stress-testing further exposed model-specific vulnerabilities in recommendation quality, which degraded by 7.4\% under amplified informational complexity. This decline contrasted with marginal improvements in rationality (+2.0\%), readability (+1.7\%) and diagnosis (+4.7\%) throughout the two-round decision-making workflow, highlighting a concerning divergence between perceived coherence and actionable clinical guidance. Our findings highlight the necessity of integrating interpretability mechanisms, such as reasoning chain visualization and real-time hallucination interception into high-risk clinical workflows. This work establishes the sequential validation framework for LLM reliability in dynamic surgical environments and advocates for a paradigm shift from accuracy-focused benchmarks to safety-aware, reasoning-grounded evaluation in medical AI.

\end{abstract}

\section{Introduction}
Large language models (LLMs) are advanced artificial intelligence systems capable of parsing and generating human-like text, demonstrating unprecedented capabilities in healthcare\cite{sallam2023, bedi2025, thirunavukarasu2023}. LLMs offer significant potential to revolutionize healthcare and facilitate clinical workflows \cite{iqbal2025, zhang2025b, wang2024}, by automating data extraction \cite{lee2025}, synthesizing medical literature\cite{croxford2025}, assisting differential diagnoses\cite{goh2024a, wu2024, liu2025, jeon2025}, interpreting radiographic or pathologic reports \cite{sallam2023}, personalizing patient care by identifying needs and preferences \cite{li2025, hunter2024}, augmenting clinician training\cite{waldock2024}, and streamlining administrative tasks \cite{moor2023}, among others.

Despite their benefits, the use of LLMs still carries inherent risks\cite{kwong2024}. Hallucinations, reported in numerous studies\cite{asgari2025, shah2024, goh2024b, wei2022, miao2024, chen2025}, are a particular concern in clinical settings. These can manifest as interpretation errors, contextual misalignment, and embedded biases that may jeopardize patient safety and trust\cite{asgari2025, shah2024, yao2025}. Studies have reported using several frameworks to assess the fidelity mismatch between LLM outputs and healthcare ground truth \cite{asgari2025, zhou2024}. This tension between potential benefit and inherent risk highlights the critical need for rigorous, context-aware evaluation frameworks before widespread deployment. However, current hallucination evaluation frameworks mainly account for the severity of errors using taxonomies, the repeatability of LLM outputs, and the potential harms of the errors\cite{asgari2025, zhou2024}.

To tackle complex reasoning tasks, modern LLMs have employed chain-of-Thoughts (CoT) prompting strategies to enhance LLMs' ability to parse, analyze, reflect upon, and ultimately solve complex problems\cite{miao2024, wei2022}. Models incorporating CoT, such as GPT-o1\cite{jaech2024} and DeepSeek-r1\cite{guo2025}, demonstrate capabilities surpassing conventional instruction-based LLMs, better mimicking human-like logical reasoning for intricate challenges\cite{jaech2024, mondillo2025, chen2025}. Much research assessing LLMs in medicine has focused on the potential of serving as diagnostic aids within the physician-AI collaborative paradigm in clinical practice \cite{asgari2025}. However, the reliance on the intermediary reasoning steps introduces novel risks, which render conventional benchmarking metrics inadequate for evaluating the clinical reliability of the outputs from thinking models with CoTs.

On the one hand, current comprehensive benchmarking frameworks for CoT LLMs did not address the clinical healthcare setting\cite{chen2025, zhang2025a}, such as peri-operative governance. For instance, benchmarking frameworks reported by Jiang et al. only focused on the LLM performances in math, science, optical character recognition (OCR), logic, space-time, and general scenes, neglecting healthcare setting \cite{zhou2024}.

On the other hand, currently reported evaluation metrics for LLM in healthcare were based on question banks to benchmark LLM performance against medical knowledge mastery within specific disease domains or clinical subspecialties\cite{zhou2024, mondillo2025, chen2025, li2024, shah2024}. Such approaches focused on static medical exams or datasets, hence failing to capture the dynamic, interdependent, and often ambiguous nature of real-world clinical reasoning\cite{asgari2025, wang2025c}. For instance, Wang et al. only focused on the use of LLM in healthcare conversations, neglecting diagnostics and treatment planning\cite{wang2025b}. However, in real-world clinical scenarios, patient symptoms are often non-specific, evidence can conflict, and disease progresses and evolves, while not all auxiliary tests are available.

Therefore, to effectively evaluate the use of LLMs with CoT in real-world peri-operative governance settings, a novel benchmarking evaluation framework is needed to assess the diagnostic precision, quality of clinical action of the LLM-proposed examination and treatment suggestions, the reasoning robustness, output structure, and knowledge alignment\cite{yao2025}.

Spinal surgeries present a uniquely compelling focus for such evaluation due to their exceptional clinical complexity and high-stakes decision-making requirements\cite{zhang2019, yoo2019, choi2017, cheriyan2015, hanada2021, mallepally2020}. Spinal disorders represent a massive disease burden worldwide, spanning from common degenerative diseases like degenerative cervical myelopathy, the leading global cause of spinal cord impairment\cite{nouri2015, witiw2017}, to rather rare and complex diseases like adolescent idiopathic scoliosis, the most common pediatric spinal disorder\cite{muhly2016, martin2014}. Moreover, low back pain alone is a primary driver of disability and the leading reason for medical consultations and abstinence from work\cite{hart1995}, with its prevalence and burden increasing with age\cite{gbd2018, hoy2012}. This confluence of high clinical prevalence, procedural complexity, and severe consequence profiles establishes spinal surgery as an optimal domain for stress-testing LLMs’ clinical reasoning capabilities and safety alignment in high-risk scenarios.

This study introduces a novel reasoning-oriented framework to assess LLMs in clinical settings, using spine surgery as a demonstration. Our primary goals are threefold:

First, we establish a sequential framework to stratify the clinical workflow timeline, providing a nuanced evaluation that integrates three critical dimensions: diagnostic logic verification, therapeutic feasibility, and safety risk anticipation. This approach moves beyond performance benchmarks to capture the dynamic, multi-stage nature of clinical reasoning.

Second, we introduce a pioneering methodology of targeted stress-testing at high-stakes surgical decision points. This is designed to rigorously probe LLM robustness under complexity, with performance quantified through a novel multi-metric assessment of diagnostic precision, clinical action quality, reasoning robustness, output coherence, and knowledge alignment.

Finally, we create a transferable deployment foundation that prioritizes interpretability through reasoning chain visualization and safety through real-time hallucination interception. This will establish standards for the integration of LLM into high-risk surgical workflows, specifically engineered to mitigate clinical errors at critical multidisciplinary coordination nodes.

\section{Results}

\subsection{Inter-Examiner Reliability and Disease-Specific Consistency}

\begin{figure}[H]
    \centering
    \IfFileExists{fig1.png}{\includegraphics[width=\textwidth]{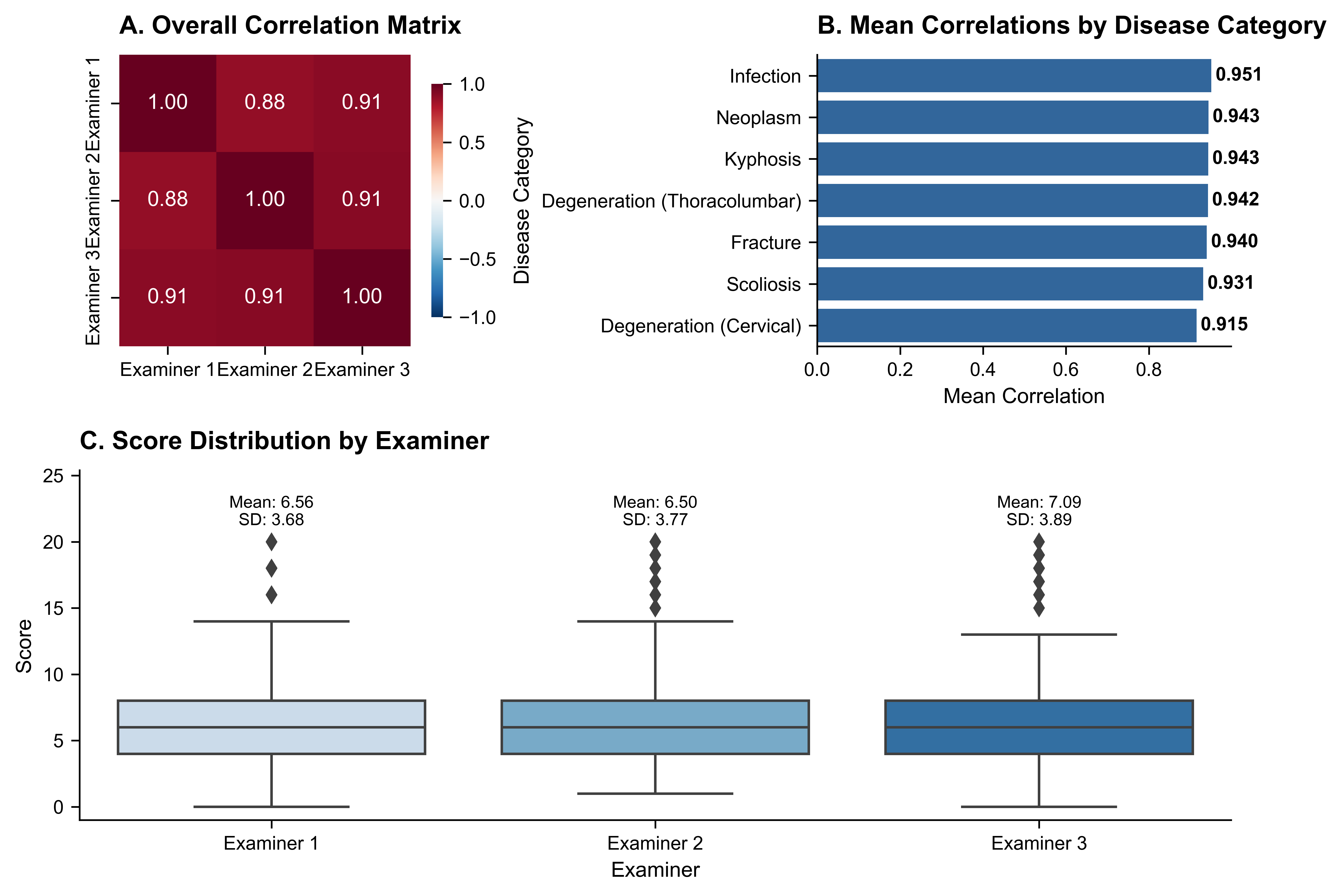}}{\fbox{fig1.png missing}}
    \caption{Agreement analysis of spinal disorder scoring among three examiners. \textbf{(a)} correlation heatmap, \textbf{(b)} disease-specific concordance, and \textbf{(c)} score distribution.}
    \label{fig:reliability}
\end{figure}

The study demonstrates robust inter-examiner reliability (r = 0.90 ± 0.014) across three independent evaluators (Examiners 1--3), as evidenced by strong pairwise correlations in the overall correlation matrix (Fig.~\ref{fig:reliability}a). Examiners 1 and 3 exhibited the highest agreement (r = 0.91), followed by Examiners 2 and 3 (r = 0.91), with Examiners 1 and 2 showing slightly lower but still substantial concordance (r = 0.88).

Disease-specific analyses revealed consistently high correlations across cervical and thoracolumbar degeneration, scoliosis and kyphosis deformities, trauma, infection and neoplasms (Fig.~\ref{fig:reliability}b). Infection test showed the strongest agreement (mean r = 0.95), while cervical degenerative diseases test demonstrated the weakest (mean r = 0.91), though all remained statistically significant. This highlights consistent performance across diverse diagnostic domains. Score distributions (Fig.~\ref{fig:reliability}c) revealed minor mean differences (Examiner 1: 6.56 ± 3.68; Examiner 2: 6.50 ± 3.77; Examiner 3: 7.09 ± 3.89), but overlapping interquartile ranges indicated comparable variability and no systematic bias, reinforcing methodological reproducibility.

\subsection{Overall Performance Across Models}

Cases cover diverse spinal pathologies and associated complexities. A staged diagnostic approach was employed: First-visit assessment (First Round) based on baseline clinical data, followed by a reassessment (Second Round) that incorporated follow-up findings and imaging results.

\begin{figure}[H]
    \centering
    \IfFileExists{fig2.png}{\includegraphics[width=0.8\textwidth]{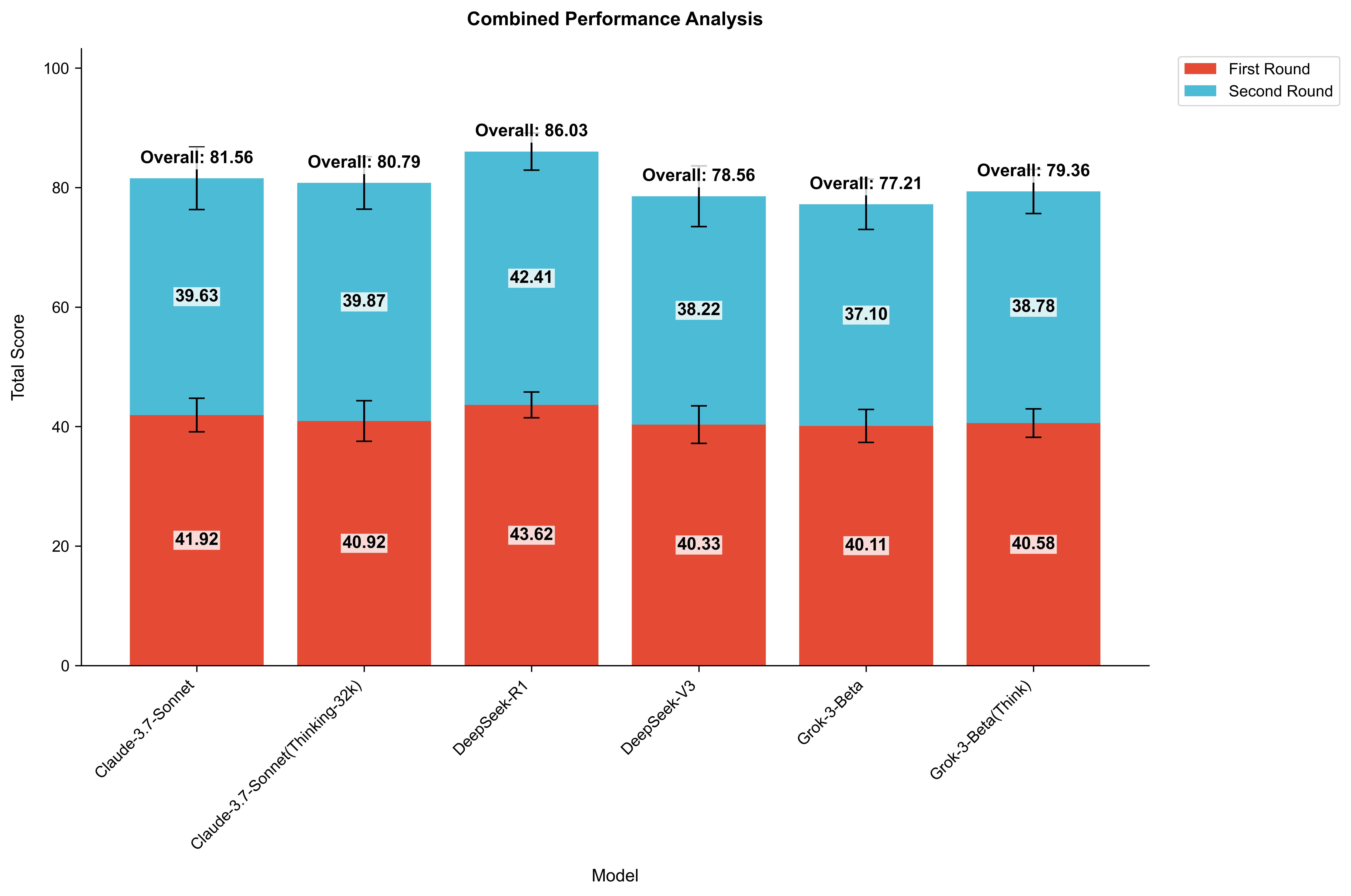}}{\fbox{fig2.png missing}} % 调整宽度以适应页面布局
    \caption{Comparative performance of large language models in two-round evaluation of spine surgical cases.}
    \label{fig:performance}
\end{figure}

The two-round evaluation revealed significant performance differences among the models (Fig.~\ref{fig:performance}). While most models' scores declined in the second round, this was attributable to an increase in task complexity and the introduction of more nuanced evaluation criteria. The elevated difficulty baseline means the decline should not be interpreted as a regression in the models' underlying capabilities.

Among the models assessed, DeepSeek-R1 demonstrated the highest overall performance with a total score of 86.03 ± 2.08, maintaining strong results across both rounds (from 43.62 ± 1.12 to 42.41 ± 1.84). Its trajectory outperformed the counterpart DeepSeek-V3, which showed more stable but less dynamic performance (Total: 78.56 ± 2.08).

Grok-3-Beta(Think) exhibited improvement over its base version. With a combined score of 79.36 ± 2.08, it significantly outperformed the standard Grok-3-Beta, which scored 77.21 ± 2.08, demonstrating that the "Think" parameter enhanced consistency and reasoning accuracy. However, Grok variants still ranked lowest among all evaluated models.

Notably, the Claude-3.7-Sonnet family was the only group in which the extended thinking mode underperformed compared to the standard configuration. The standard variant achieved a total score of 81.56 ± 1.92, surpassing the extended thinking version at 80.79 ± 1.83. This suggests that increasing inference-time thinking capacity did not yield measurable benefits for this model family.

\subsection{Disease-Specific Performance Variability}

\begin{figure}[H]
    \centering
    \IfFileExists{fig3.pdf}{\includegraphics[width=\textwidth]{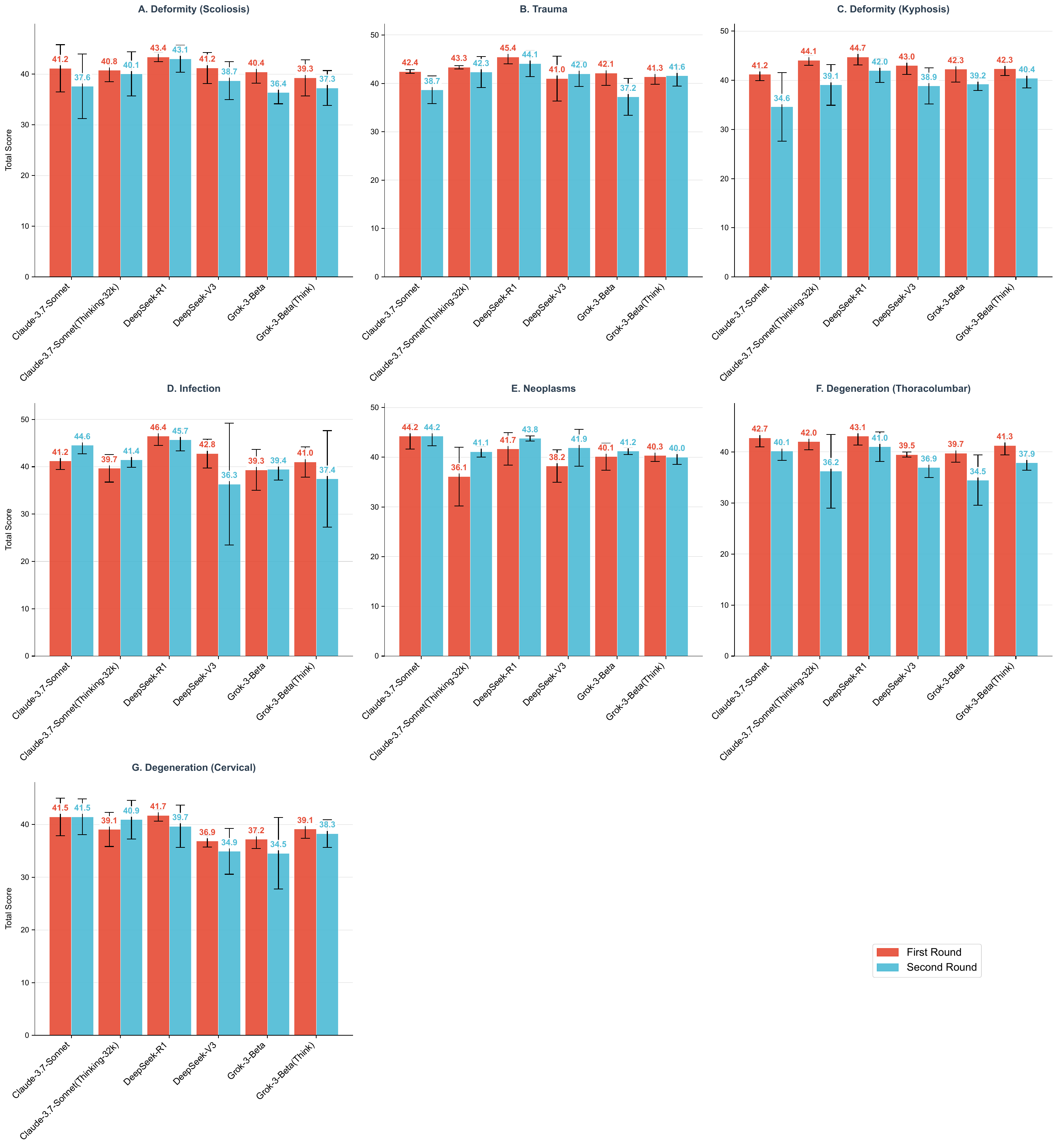}}{\fbox{fig3.pdf missing}} % 调整宽度以适应页面布局
    \caption{Two-round model performances across spinal disease categories. \textbf{(a)} Deformity (Scoliosis). \textbf{(b)} Trauma. \textbf{(c)} Deformity (Kyphosis). \textbf{(d)} Infection. \textbf{(e)} Neoplasms. \textbf{(f)} Degeneration (Thoracolumbar). \textbf{(g)} Degeneration (Cervical).}
    \label{fig:disease_performance}
\end{figure}

The multi-disease evaluation across seven clinical categories, including cervical and thoracolumbar degeneration, scoliosis and kyphosis deformities, trauma, infection and neoplasms revealed distinct patterns of model performance and adaptation. While most models exhibited resilience to the heightened task complexity introduced in the second round, the magnitude of improvement or decline varied significantly according to disease domain and model architecture (Fig.~\ref{fig:disease_performance}a--g).

A key finding was DeepSeek-R1's consistent superiority over DeepSeek-V3 in high-stakes clinical domains, particularly trauma and infection. For trauma, DeepSeek-R1 maintained a leading score of 45.44 ± 0.96 (Round 1) and 44.11 ± 1.28 (Round 2). Similarly, regarding infection, it achieved 46.44 ± 1.33 and 45.67 ± 1.57 across rounds, demonstrating robust and adaptable inference capabilities despite increasing evaluation complexity.

Across scoliosis, trauma, and kyphosis evaluations, Claude-3.7-Sonnet consistently outperformed the Claude-3.7-Sonnet (Thinking:32k) counterpart. These results suggest that the additional thinking capacity did not yield measurable improvements in these areas, making Claude the only model family where the "thinking" variant underperformed relative to its standard version. Parallel trends emerged in Grok variants: "thinking" mode produced no significant improvements in scoliosis, trauma, infection, or neoplasms sections. This persistent capability gap emphasizes architectural limitations and distances these models from top-tier benchmarks.

\subsection{Where and When Hallucinations Thrive - Multidimensional Evaluation}

\begin{figure}[H]
    \centering
    \IfFileExists{fig4.jpg}{\includegraphics[width=\textwidth]{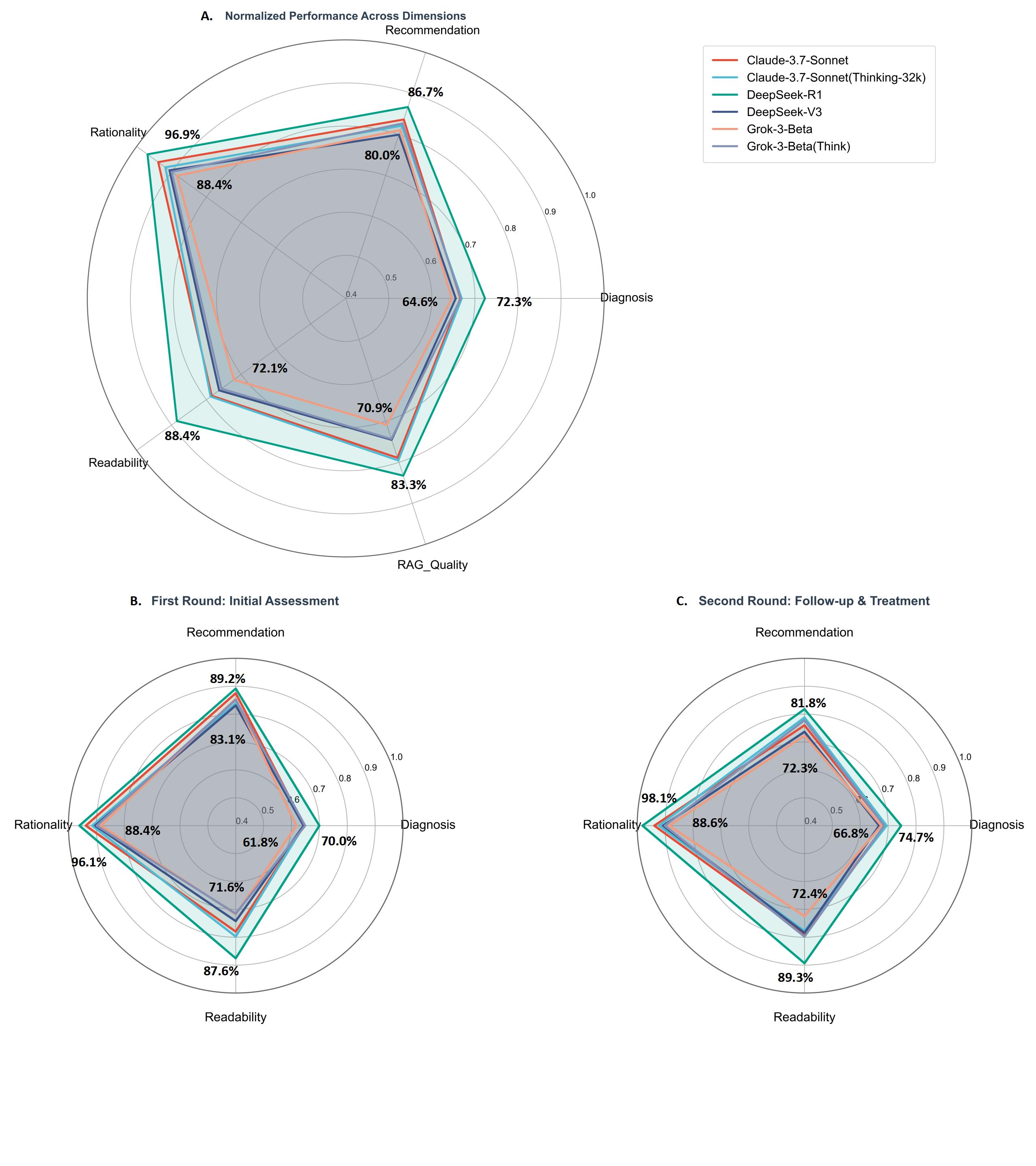}}{\fbox{fig4.jpg missing}}
    \caption{Comparative assessment of large language models across clinical-oriented dimensions in spinal surgical decision support. \textbf{(a)} Overall performances. \textbf{(b)} First-Round initial assessment. \textbf{(c)} Second-Round follow-up \& treatment assessment.}
    \label{fig:multidimensional_eval}
\end{figure}

Comprehensive multidimensional evaluation revealed critical engineering insights through cross-dimensional stress testing (Fig.~\ref{fig:multidimensional_eval}a), with DeepSeek-R1 demonstrating superior stability and peak capabilities across clinical reasoning domains. Crucially, the RAG quality, fundamental for hallucination suppression during inference, showed significant divergence (DeepSeek-R1: 83.3\% vs. Grok-3-Beta: 70.9\%). While DeepSeek-R1 maintained consistently high performance across all dimensions, diagnostic precision emerged as critical vulnerability, with even peak performance (DeepSeek-R1: 72.3\%) falling below clinical deployment thresholds for high-stakes spinal decision support.

Fig.~\ref{fig:multidimensional_eval}b and~\ref{fig:multidimensional_eval}c revealed DeepSeek-R1's adaptive superiority in dynamic clinical workflows but exposes a critical vulnerability in recommendation stability. While the model maintained high cross-dimensional robustness (e.g., from 87.6\% to 89.3\% by RAG quality), recommendation performance decreased by 7.4\%. This divergence contrasts with marginal gains in rationality (+2.0\%), readability (+1.7\%) and diagnosis (+4.7\%). Crucially, recommendation accuracy, which is highly dependent on retrieved evidence, appeared disproportionately sensitive to the amplified informational complexity characteristic of second-round scenarios.

This trend carries significant clinical implications: elevated rationality (98.1\%) coinciding with compromised recommendation quality (81.0\%) may produce deceptively coherent but clinically unsound outputs. Such "rational hallucinations" introduce critical safety risks in spinal decision support contexts, as clinicians struggle to identify plausible, yet erroneous recommendations grounded in seemingly logical reasoning. These observations underscore the necessity of workflow-stress testing beyond aggregate metrics to expose context-dependent failure modes.

\subsection{Complexity-Dependent Performance Across AI Stress Test}

\begin{figure}[H]
    \centering
    \IfFileExists{fig5.jpg}{\includegraphics[width=\textwidth]{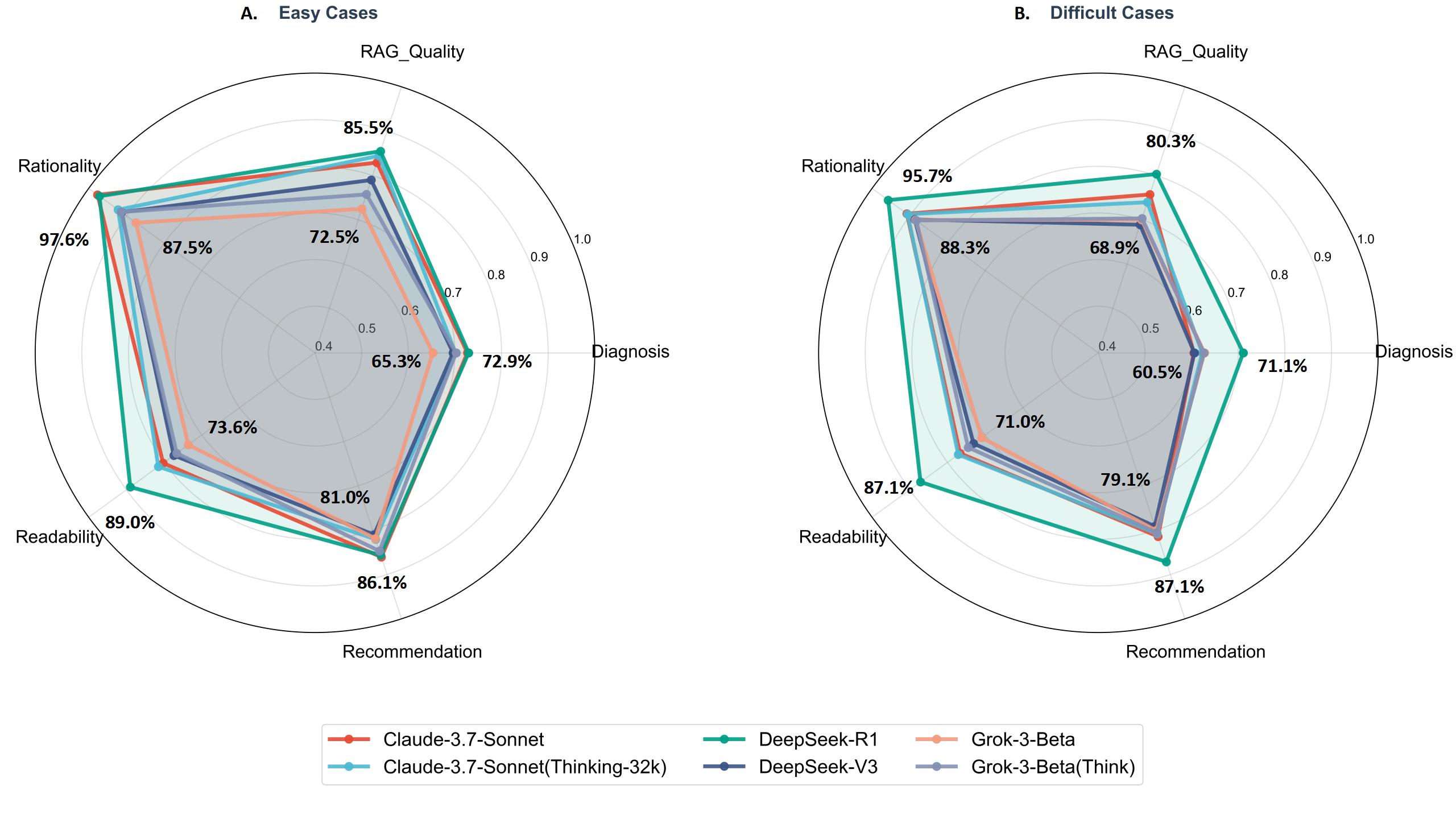}}{\fbox{fig5.jpg missing}}
    \caption{AI-oriented comparative performance of large language models for spinal surgical diseases. \textbf{(a)} easy clinical cases. \textbf{(b)} difficult clinical cases.}
    \label{fig:complexity_test}
\end{figure}

Comprehensive evaluation across dual difficulty tiers revealed critical deployment limitations in current clinical AI systems. In easy cases (Fig.~\ref{fig:complexity_test}a), Claude-3.7-Sonnet demonstrated selective superiority in rationality (97.6\%) and recommendation (86.1\%), whereas DeepSeek-R1 dominated other metrics, achieving peak performance in RAG quality (85.5\%), readability (89.0\%) and diagnosis (72.9\%). Under difficult case assessment (Fig.~\ref{fig:complexity_test}b), DeepSeek-R1 exhibited exceptional resilience, consistently outperforming all competitors with leading scores in rationality (95.7\%), recommendation (87.1\%), RAG quality (80.3\%), diagnosis (71.1\%), and readability (87.1\%). Performance scaling validated expected competency gaps between tiers, with RAG quality demonstrating the most significant degradation (5.2\% absolute reduction). Critically, diagnosis performance across models in difficult cases and RAG quality fell below clinically acceptable thresholds for high-stakes spinal decision support, indicating unresolved reliability barriers in complex scenarios.

\section{Discussion}

This study demonstrates that while experienced physicians rated the performance of advanced LLMs, our clinician-centered framework reveals a more critical finding: even the most capable models used remain unsuitable for autonomous deployment due to persistent and high-stakes reasoning errors. Superior performance underscores the value of reasoning models, yet no model consistently met the safety thresholds required for unsupervised use in the workflow of spinal surgery. These results align with prior studies highlighting LLMs' propensity for clinically hazardous inaccuracies\cite{asgari2025, shah2024, yao2025} and their limited robustness in real-world medical contexts\cite{asgari2025, wang2025c}. This reinforces that LLMs are best positioned as decision-support tools under expert clinician oversight, particularly for tasks like differential diagnosis where their outputs align with established paradigms.

The primary merit of this work lies in its novel framework, which leverages deep clinical expertise to move beyond accuracy metrics toward evaluating reasoning fidelity. By stratifying the clinical workflow and embedding high-stakes decision points validated by spinal surgeons, our approach provides a more authentic stress test of model robustness. This approach successfully exposes critical limitations, lacking dynamic guideline alignment, complexity-sensitive reasoning, and clinical safety, that are invisible to simpler benchmarks. Our framework is designed for locating hallucination, which is defined in AI engineering as the production of fluent yet factually inaccurate content with low per-token likelihood, but which is evaluated in the clinical realm solely by its capacity to harm patients through misinformation, irrespective of statistical credibility\cite{asgari2025, yao2025, lanham2023}. Concretely, we capture instances where an autoregressive LLM ventures beyond its training manifold and emits assertions that contradict authoritative medical knowledge or the supplied context\cite{yao2025}. While extant studies have sporadically documented hallucinations within discrete phases of healthcare\cite{yao2025, kwong2024}, our methodology provides a comprehensive lens to evaluate this risk across the entire peri-operative continuum by encompassing pre-operative evaluation, intra-operative decision-making, and post-operative surveillance, which remains empirically under-characterized in real clinical populations.

Hallucination management mitigating hallucination-induced uncertainty demands a multi-layered, peri-operative governance framework that embeds safeguards at every point where generative models touch spinal care\cite{yao2025}. First, at the data layer, curate continuously updated, subspecialty-specific corpora by integrating operative notes, intra-operative imaging, and post-operative outcomes to narrow the gap between training distribution and real-world variability. Second, at the inference layer, implement retrieval-augmented generation that anchors each token prediction to authoritative ontologies and enforces temporal consistency checks against longitudinal EHR timelines. Third, at the human-AI interface, surgeon-in-the-loop confirmation for any model-generated diagnosis, implant sizing, or complication forecast, supported by visual saliency maps that highlight the imaging voxels or text spans underpinning each recommendation. Fourth, institute post-deployment surveillance: federated error-tracking dashboards that log discrepancies between predicted and observed events across institutions, enabling rapid retraining and bias correction. Finally, cultivate AI-literate surgical teams through scenario-based simulation training and transparent model-cards that clarify limitations and failure modes. Only through this iterative, evidence-driven stewardship can healthcare workers harness generative AI without compromising patient safety or professional accountability.

It is essential to recognize, however, that such architectural safeguards are necessary but insufficient without disciplined model training. Reasoning models are not innate solutions to hallucination; their propensity to fabricate is tightly coupled to their training regimen. Recent evidence demonstrates that models subjected to the full curriculum, supervised fine-tuning followed by reinforcement learning, exhibit markedly fewer hallucinations, whereas those distilled or tuned with reinforcement learning alone succumb to higher fabrication rates\cite{lanham2023}. For AI engineers, the implication is unambiguous: adopt the complete training pipeline as a non-negotiable baseline and reserve single-stage shortcuts only for low-risk, well-monitored applications. Continuous post-deployment auditing should then track hallucination incidence across real clinical encounters, enabling iterative retraining that keeps the model's distribution anchored to authoritative, subspecialty-curated data.

Moreover, while Chain-of-Thought (CoT) prompting enhances the interpretability of AI decision-support systems by elucidating the reasoning process, our findings underscore that it concurrently introduces a distinct class of hallucination risks. The very length and complexity of the reasoning chain, which is intended to mirror clinical cognition, can become a source of error propagation and confabulation. Specifically, (1) Error Accumulation: A minor inaccuracy or unwarranted assumption in an early step of the chain can be amplified through subsequent inferences, leading to a logically structured but fundamentally flawed conclusion\cite{lanham2023}. (2) The Illusion of Validity: Elaborate, coherent reasoning can create a persuasive "narrative fallacy," where the sheer volume of plausible-sounding text may obscure underlying factual inaccuracies or gaps in evidence, making it more challenging for clinicians to detect hallucinations than in concise, output-only responses\cite{barez2025}. (3) The Opacity of Complexity: Paradoxically, a lengthy CoT can be as difficult to critically evaluate as a "black box" model, as clinicians may lack the time or resources to verify every logical step in practice\cite{barez2025, shojaee2025}. Our sequential validation framework is specifically designed to mitigate these risks by deconstructing the CoT into distinct, verifiable stages, such as initial diagnosis, investigation rationale, and treatment planning. Thereby, subjecting each segment of the reasoning process to independent scrutiny. This approach not only helps in localizing where hallucinations originate within the chain but also provides a methodological safeguard against the seductive but potentially misleading coherence of AI-generated rationales.

This study has several important limitations that also delineate the scope of our contribution. As a pilot investigation, the sample size from a single institution is limited, which restricts the statistical power for subgroup analyses and broad generalizability. The exclusive use of a simplified Chinese-language case series, while providing a focused validation context, may introduce confoundings related to variable model pre-training and limit the immediate applicability of findings to other linguistic and clinical settings. Furthermore, the evaluation is restricted to textual reasoning without incorporating radiographic images, which is a critical limitation for a radiographically-driven specialty like spine surgery, impacting the construct validity of the diagnostic assessment. These design choices reflect the pilot nature of this work, which aims to propose and preliminarily validate a framework rather than deliver definitive efficacy evidence.

By foregrounding clinical expertise in its design, this framework establishes a new standard for the translational validation of LLMs, positioning them not as autonomous oracles but as accountable, evidence-bound partners in care.

\section{Conclusion}

This work proposed a clinician-centred, multidimensional framework that reorients the evaluation of large language models (LLMs) in surgery from narrow accuracy benchmarks toward a holistic assessment of reasoning fidelity, retrieval grounding, and pathology-specific robustness. By stratifying the clinical workflow timeline, this study provides the first replicable standard for evaluating LLMs under dynamically complex, high-stakes scenarios.

While a leading model consistently outperformed its peers in our Chinese case bank, demonstrating superior diagnostic precision and therapeutic appropriateness, even the best-performing model exhibited significant performance decay in ambiguous contexts. This critical finding underscores that extended chain-of-thought traces are not a reliable proxy for clinical safety.

Our findings necessitate a paradigm shift in how LLMs are validated for clinical use. First, diagnostic accuracy alone is an insufficient benchmark. Dynamic alignment with evolving guidelines and strict grounding in subspecialty ontologies are essential to prevent hazardous deviations. Second, interpretability requires standardization: our clinician-rated chain-of-thought checklist offers a validated, deployable safety screen for high-stakes workflows. 

This framework, detailed for replication in Fig~\ref{fig:7}, establishes a new reference standard for the translational validation of LLMs. By foregrounding reasoning traceability and workflow-integrated robustness, it positions these tools not as opaque oracles, but as accountable, evidence-bound partners in patient care. We encourage its adoption beyond spinal surgery to ensure the rigorous and safe deployment of AI across all surgical disciplines.

\section{Methods}

\begin{figure}[H]
    \centering
    \IfFileExists{fig6.jpg}{\includegraphics[width=\textwidth]{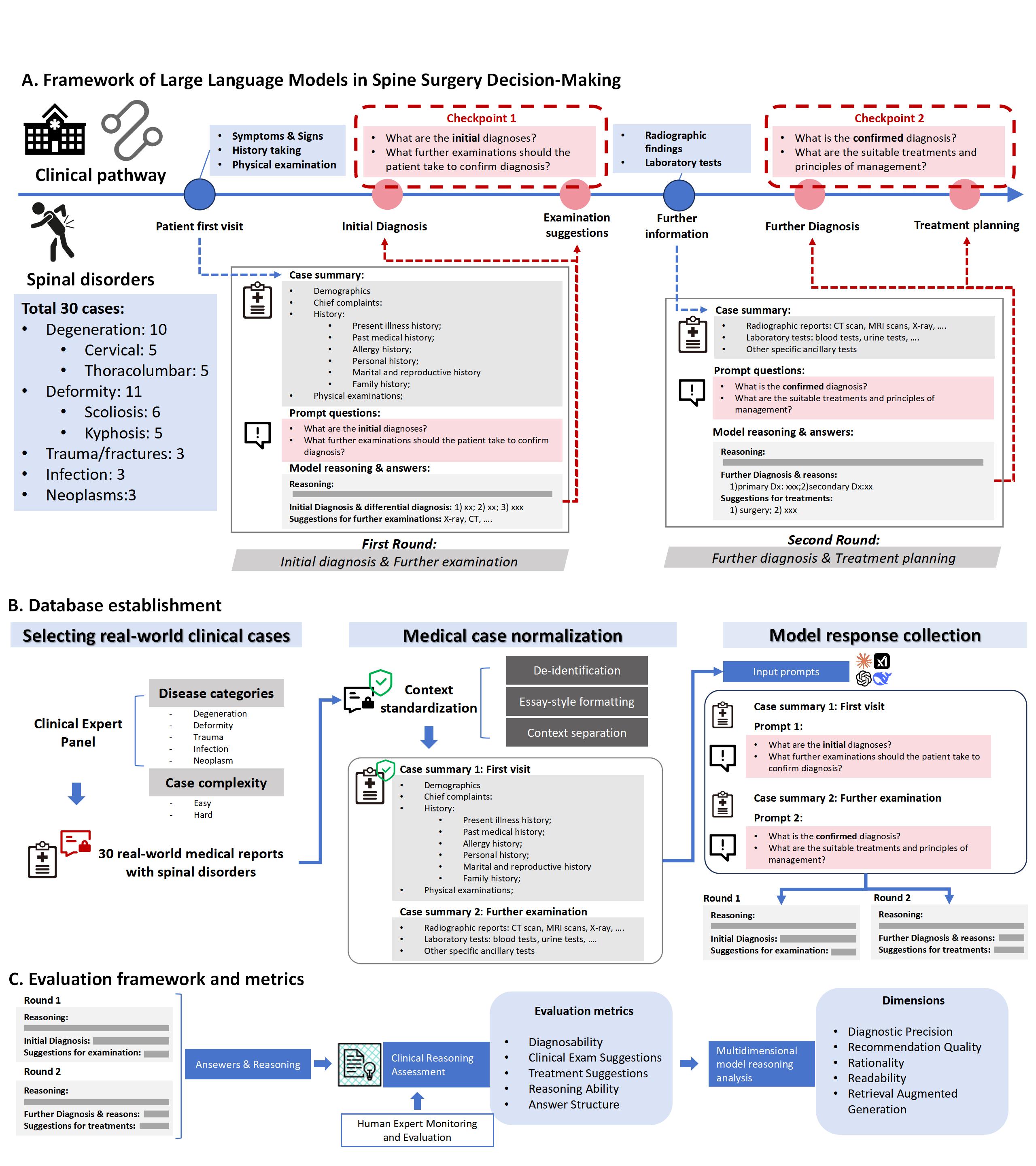}}{\fbox{fig6.jpg missing}} % Please ensure the image file exists
    \caption{Framework integrating large language models (LLMs) for spine surgical decision-making and clinical evaluation. \textbf{(A)} Diagnostic and therapeutic workflow of LLMs in spinal disorder management, demonstrating a two-stage clinical pathway. \textbf{(B)} Database establishment including case selection and normalization, model response collection. \textbf{(C)} The procedure of multidimensional evaluation and analyses of model response.}
    \label{fig:6}
\end{figure}

To rigorously assess the efficacy of both instruction-based and reasoning-oriented LLMs in clinical practice, we developed a comprehensive evaluation framework (Fig.~\ref{fig:6}). Our methodology involved: 

\begin{enumerate}
    \item Curating real patient cases from clinical practice;
    \item Standardizing these cases using a unified prompt structure to facilitate LLM reasoning and response generation; and
    \item Systematically collecting both intermediate reasoning steps and final outputs.
\end{enumerate}
These LLM-generated responses were then evaluated against novel assessment metrics by a panel of three experienced spine surgeons.

\subsection{Data Preparation}

To enable targeted stress testing of the LLMs, we curated a library of 30 expert-validated spinal cases spanning the five major pathological domains. This purposive sampling strategy prioritizes clinical depth and complexity over epidemiological breadth, with each domain designed to probe specific, high-stakes reasoning failures: Degenerative diseases (10 cases) assess stepwise stenosis/disc herniation management; deformity cases (11) evaluate Cobb angles, osteotomies, and comorbidity management; trauma (3) focus on fracture classification and emergent interventions; infection (3) probe antimicrobial stewardship; neoplasms (3) examine metastatic strategies and surgical margins.

A panel of clinical experts reviewed and validated the selected cases (presented in simplified Chinese) to ensure their clinical relevance and quality. These cases were derived from a comprehensive database of real-world spinal surgery cases compiled in three phases. These cases represent diverse spinal pathologies and procedural complexities, covering cervical and thoracolumbar degeneration, scoliosis and kyphosis deformities, trauma, infection and neoplasms. All patient data has been rigorously anonymized and appropriately structured to safeguard privacy while preserving clinical integrity and relevance.

Clinical cases were sourced from real-world medical records in the first phase. The focus was on capturing initial visit diagnoses and inpatient treatment plans. Cases were selected to ensure clear diagnostic and therapeutic contexts, representing diverse clinical scenarios applicable to spine department practice. The case materials include the patient's chief complaint, present illness history, and physical examination findings, along with relevant laboratory and diagnostic medical reports. Regrettably, due to the currently limited capability of Large Language Models (LLMs) in interpreting medical imaging, we have extracted the textual information directly from the medical reports. This includes imaging descriptions and preliminary diagnoses.

The second phase is to transform case narratives into standardized essay-style prompts designed to evaluate clinical reasoning and judgment. This involved reformatting descriptive contexts into structured inputs while rigorously removing all identifiable information to ensure complete anonymization without compromising essential clinical details.

Generated LLM responses and reasoning traces underwent evaluation in the final phase. A structured, multidimensional assessment framework was implemented to score responses systematically.

\subsection{Disease Category and Complexity}

All cases were derived from real patient admissions to the Spinal Surgery Department of The University of Hong Kong Shenzhen Hospital between March 2024 and March 2025. A random sample of 30 cases was selected for classification and complexity assessment by an expert panel. Cases were categorized using classical spinal surgery etiological principles, comprehensively covering major spinal pathologies including degenerative diseases, deformity, trauma, infection, and neoplasms. Complexity assessment defined simple cases as those meeting all criteria: single etiology, typical symptoms and signs, and characteristic imaging findings. Complex cases were defined by the presence of any of the following features: multiple etiologies or severe comorbidities, rare disease, atypical symptoms and signs, or imaging findings presenting additional diagnostic or therapeutic challenges.

\subsection{Prompting Formation}

The question bank employs a rigorous prompting framework designed to systematically evaluate clinical reasoning by simulating authentic diagnostic workflows through a structured two-phase approach. All cases adhere to a standardized architecture that sequentially presents information: chief complaint, history, physical examination findings, and investigations. Phase I mimics the initial patient encounter, providing only the chief complaint, history, and examination findings. This phase assesses foundational diagnostic reasoning by requiring respondents to generate differential diagnoses ranked by probability and propose appropriate next-step investigations. Phase II simulates the comprehensive clinical scenario after all data is available, incorporating actual investigative reports (laboratory, imaging). This phase evaluates advanced clinical judgment by demanding updated differential diagnoses, exposition of disease-specific treatment principles incorporating the latest evidence-based advancements, and formulation of individualized management plans for the presented case. Across both phases, therapeutic planning mandates precise specification, including hierarchical conservative management sequences and, where applicable, anatomically precise surgical strategies detailing approach, levels, and instrumentation. This progressive framework compels respondents to dynamically build clinical logic, moving from initial hypothesis generation through to definitive management planning. (Refer to Supplementary Note 1 for a full case narrative exemplifying this structure).

\subsection{Model Selection}

To facilitate a comparative analysis of standard instruction-following models versus their reasoning-optimized versions, six models were selected: Deepseek-V3, Deepseek-R1, Grok-3-beta, Grok-3-beta (Think), Claude 3.7 Sonnet, and Claude 3.7 Thinking. Specifically, this allows pairwise comparisons between Deepseek-V3 and Deepseek-R1, Grok-3-beta and Grok-3-beta (Think), and Claude 3.7 Sonnet and Claude 3.7 Thinking.

\subsection{Scoring Framework for Clinical Reasoning Assessment}

\begin{figure}[H]
    \centering
    \IfFileExists{fig7.jpg}{\includegraphics[width=\textwidth]{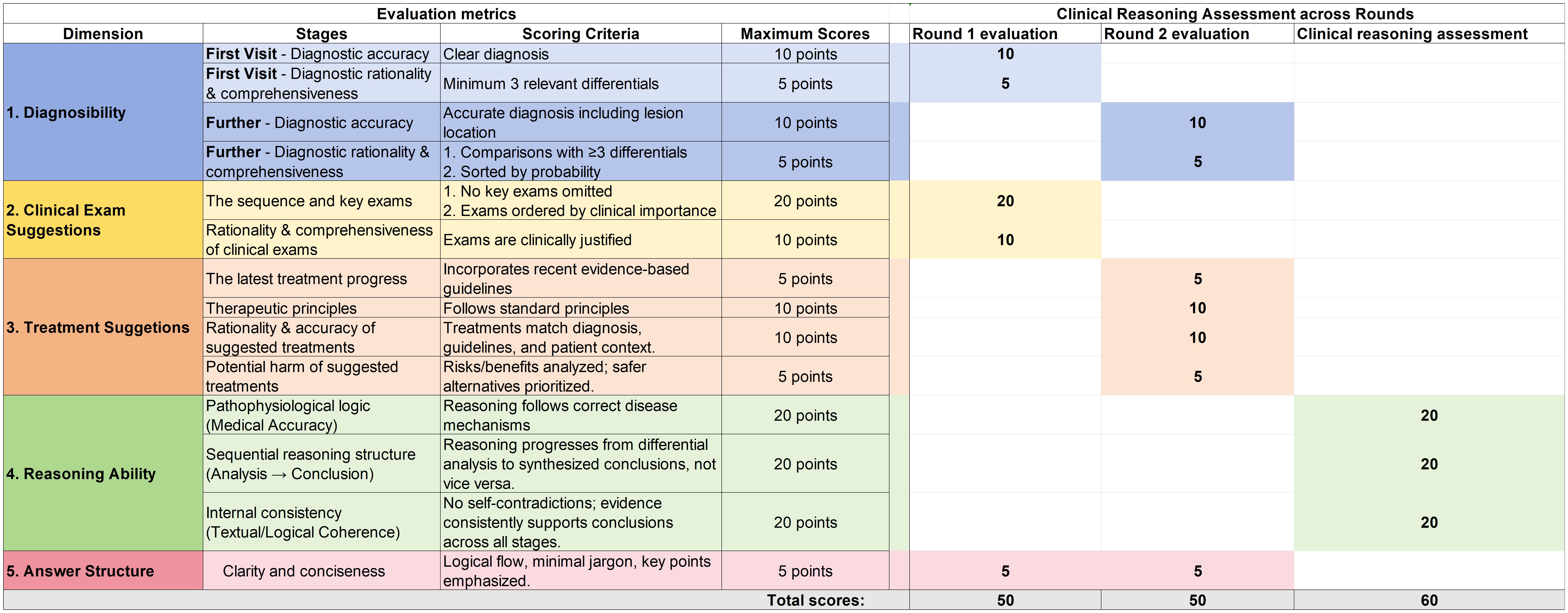}}{\fbox{fig7.jpg missing}} % Please ensure the image file exists
    \caption{Multidimensional Clinical Reasoning Evaluation Framework for Spine Surgery Decision Support Systems.}
    \label{fig:7}
\end{figure}

As detailed in Fig~\ref{fig:7}, clinical reasoning was assessed across five weighted domains: First-visit Diagnosability (max 30 points), Clinical Exam Suggestions (max 30), Treatment Suggestions (max 30), Reasoning Ability (max 60) and Answer Structure (max 10). Domain-specific deductions identified reasoning and generating flaws (e.g., unsorted differentials, omitted key examinations). This granular scoring system provided quantitative proficiency metrics while highlighting targets for educational remediation.

\subsection{AI-Oriented Multidimensional Performance}

To complement the clinical scoring framework and provide AI-specific performance insights, we conducted separate multidimensional analysis focused on model reasoning characteristics by regrouping evaluated items in Fig~\ref{fig:7}. This analysis examined five key dimensions: 

\begin{enumerate}
    \item \textbf{Diagnostic precision}, evaluating first-visit and further diagnostic accuracy;
    \item \textbf{Suggestions quality of clinical action}, assessing examination sequencing and treatment suggestion;
    \item \textbf{Reasoning robustness}, measuring diagnostic thoroughness and treatment safety considerations;
    \item \textbf{Output structure}, analyzing logical coherence and presentation clarity; and
    \item \textbf{Knowledge alignment}, verifying adherence to current guidelines through RAG performance evaluation.
\end{enumerate}

This orthogonal assessment approach provides researchers with granular signals about model behavior patterns, error typologies, and optimization priorities distinct from traditional clinical proficiency metrics.

Specifically, they are Diagnostic Precision, Recommendation Quality, Rationality, Readability and Retrieval Augmented Generation (RAG) quality. Diagnostic Precision was evaluated by measuring accuracy in initial and follow-up diagnoses against clinical gold standards. Recommendation Quality assessed the clinical relevance and prioritization of examination suggestions and treatment plans based on guideline adherence and expert validation. Rationality was measured through a comprehensive evaluation of clinical reasoning logic, differential diagnosis completeness, and safety considerations in decision-making pathways. Readability quantified the structural organization, terminology precision, and narrative coherence of AI-generated reports against clinical documentation standards. RAG Quality evaluated evidence integration capabilities through assessment of medical knowledge currency and treatment guideline adherence. All metrics were scored through independent expert consensus ratings using standardized rubrics, with case-specific adjustments for pathological complexity. The weighting scheme allocated 20 points to Diagnostic Precision, 30 to Recommendation Quality, 25 to Rationality, 10 to Readability, and 15 to RAG Quality, yielding a comprehensive 100-point evaluation system that mirrors real-world clinical workflow while ensuring robust assessment of AI utility in spinal surgical care.

\subsection{Ethical Approval}

This study was approved by the University of Hong Kong - Shenzhen Hospital Review Board (Approval number: hkusz2025078). De-identified clinical records were used under institutional guidelines for retrospective studies. Informed consent was obtained from all individual participants included in the study.

\subsection{Statistical Analysis}

Statistical analyses were performed using Python (version 3.9). Packages including SciPy, pandas, scikit-learn were used to evaluate model performance and inter-rater reliability. Inter-examiner agreement was quantified through Pearson correlation coefficients for pairwise comparisons between three independent evaluators. Model comparisons were conducted using descriptive statistics (mean ± SD).

\section*{Data Availability}
The datasets generated and/or analyzed during the current study are not publicly available due to the sensitive nature of the real clinical cases involved, which contain confidential patient information and personal health data. However, they are available from the corresponding author upon reasonable request.

\section*{Acknowledgements}
The authors sincerely thank The University of Hong Kong - Shenzhen Hospital for their invaluable contributions to this study.

\section*{Funding}
This work was supported by the Sanming Project of Medicine in Shenzhen, China (Grant No. SZSM202211004); the Shenzhen Science and Technology Program (Grant No. KJZD20240903102759061); and the Shenzhen-Hong Kong Cooperation Zone for Technology and Innovation (Grant No. HZQSWS-KCCYB-2024055). The funders played no role in the study design, data collection, analysis, interpretation of data, or the writing of this manuscript.

\section*{Author Contributions}
% Using itemize for clearer presentation of individual contributions
\begin{itemize}
    \item[] \textbf{Dong Chen (D.C.)}: Analyzed and interpreted the scoring data, and was a major contributor in writing the manuscript.
    \item[] \textbf{Yanzhe Wei (Y.W.)}, \textbf{Guan-Ming Kuang (G.K.)}, \textbf{Canhua Ye (C.Y.)}, \textbf{Huiren Tao (H.T.)}: Designed the scoring framework and marked the answers generated by LLMs.
    \item[] \textbf{Zonglin He (Z.H.)}, \textbf{Meiru An (M.A.)}, \textbf{Huili Peng (H.P.)}, \textbf{Yong Hu (Y.H.)}: Revised the manuscript and figures.
    \item[] \textbf{Kenneth MC Cheung (K.C.)}: Organized and supervised the study.
    \item[] All authors read and approved the final manuscript.
\end{itemize}

\section*{Competing Interests}
The authors declare no competing interests.

\section*{Correspondence}
Correspondence and requests for materials should be addressed to Dong Chen.

\bibliographystyle{unsrt}
\bibliography{reference}

@article{sallam2023,
  author = {Sallam, M.},
  title = {ChatGPT Utility in Healthcare Education, Research, and Practice: Systematic Review on the Promising Perspectives and Valid Concerns},
  journal = {Healthcare (Basel)},
  year = {2023},
  volume = {11},
  doi = {10.3390/healthcare11060887}
}

@article{bedi2025,
  author = {Bedi, S. and others},
  title = {Testing and Evaluation of Health Care Applications of Large Language Models: A Systematic Review},
  journal = {JAMA},
  year = {2025},
  volume = {333},
  pages = {319--328},
  doi = {10.1001/jama.2024.21700}
}

@article{thirunavukarasu2023,
  author = {Thirunavukarasu, A. J. and others},
  title = {Large language models in medicine},
  journal = {Nature Medicine},
  year = {2023},
  volume = {29},
  pages = {1930--1940},
  doi = {10.1038/s41591-023-02448-8}
}

@article{iqbal2025,
  author = {Iqbal, U. and others},
  title = {Impact of large language model (ChatGPT) in healthcare: an umbrella review and evidence synthesis},
  journal = {Journal of Biomedical Science},
  year = {2025},
  volume = {32},
  pages = {45},
  doi = {10.1186/s12929-025-01131-z}
}

@article{zhang2025b,
  author = {Zhang, L. and others},
  title = {Application of large language models in healthcare: A bibliometric analysis},
  journal = {Digital Health},
  year = {2025},
  volume = {11},
  pages = {20552076251324444},
  doi = {10.1177/20552076251324444}
}

@article{wang2024,
  author = {Wang, D. and Zhang, S.},
  title = {Large language models in medical and healthcare fields: applications, advances, and challenges},
  journal = {Artificial Intelligence Review},
  year = {2024},
  volume = {57},
  pages = {299},
  doi = {10.1007/s10462-024-10921-0}
}

@article{lee2025,
  author = {Lee, D. and others},
  title = {Using Large Language Models to Automate Data Extraction From Surgical Pathology Reports: Retrospective Cohort Study},
  journal = {JMIR Formative Research},
  year = {2025},
  volume = {9},
  pages = {e64544},
  doi = {10.2196/64544}
}

@article{croxford2025,
  author = {Croxford, E. and others},
  title = {Current and future state of evaluation of large language models for medical summarization tasks},
  journal = {NPJ Health Systems},
  year = {2025},
  volume = {2},
  doi = {10.1038/s44401-024-00011-2}
}

@article{goh2024a,
  author = {Goh, E. and others},
  title = {Large Language Model Influence on Diagnostic Reasoning: A Randomized Clinical Trial},
  journal = {JAMA Network Open},
  year = {2024},
  volume = {7},
  pages = {e2440969},
  doi = {10.1001/jamanetworkopen.2024.40969}
}

@article{wu2024,
  author = {Wu, D. and Nie, L. and Mumtaz, R. A. and Agarwal, K.},
  title = {A LLM-Based Hybrid-Transformer Diagnosis System in Healthcare},
  journal = {IEEE Journal of Biomedical and Health Informatics},
  year = {2024},
  doi = {10.1109/jbhi.2024.3481412}
}

@article{liu2025,
  author = {Liu, X. and others},
  title = {A generalist medical language model for disease diagnosis assistance},
  journal = {Nature Medicine},
  year = {2025},
  volume = {31},
  pages = {932--942},
  doi = {10.1038/s41591-024-03416-6}
}

@article{jeon2025,
  author = {Jeon, S. and Kim, H.-G.},
  title = {A comparative evaluation of chain-of-thought-based prompt engineering techniques for medical question answering},
  journal = {Computers in Biology and Medicine},
  year = {2025},
  volume = {196},
  pages = {110614},
  doi = {10.1016/j.compbiomed.2025.110614}
}

@article{li2025,
  author = {Li, J. and others},
  title = {Identifying healthcare needs with patient experience reviews using ChatGPT},
  journal = {PLoS ONE},
  year = {2025},
  volume = {20},
  pages = {e0313442},
  doi = {10.1371/journal.pone.0313442}
}

@article{hunter2024,
  author = {Hunter, J. and Nicandri, G. and Bozic, K. J.},
  title = {Value-based Healthcare: How Can Large Language Model (LLM) Technology be Integrated With Patient-reported Outcomes?},
  journal = {Clinical Orthopaedics and Related Research},
  year = {2024},
  volume = {482},
  pages = {1948--1950},
  doi = {10.1097/corr.0000000000003261}
}

@article{waldock2024,
  author = {Waldock, W. J. and others},
  title = {The Accuracy and Capability of Artificial Intelligence Solutions in Health Care Examinations and Certificates: Systematic Review and Meta-Analysis},
  journal = {Journal of Medical Internet Research},
  year = {2024},
  volume = {26},
  pages = {e56532},
  doi = {10.2196/56532}
}

@article{moor2023,
  author = {Moor, M. and others},
  title = {Foundation models for generalist medical artificial intelligence},
  journal = {Nature},
  year = {2023},
  volume = {616},
  pages = {259--265}
}

@article{kwong2024,
  author = {Kwong, J. C. C. and Wang, S. C. Y. and Nickel, G. C. and Cacciamani, G. E. and Kvedar, J. C.},
  title = {The long but necessary road to responsible use of large language models in healthcare research},
  journal = {NPJ Digital Medicine},
  year = {2024},
  volume = {7},
  pages = {177},
  doi = {10.1038/s41746-024-01180-y}
}

@article{asgari2025,
  author = {Asgari, E. and others},
  title = {A framework to assess clinical safety and hallucination rates of LLMs for medical text summarisation},
  journal = {NPJ Digital Medicine},
  year = {2025},
  volume = {8},
  pages = {274},
  doi = {10.1038/s41746-025-01670-7}
}

@misc{zhou2024,
  author = {Zhou, Y. and Liu, X. and Ning, C. and Zhang, X. and Wu, J.},
  title = {Reliable and diverse evaluation of LLM medical knowledge mastery},
  howpublished = {arXiv preprint arXiv:2409.14302},
  year = {2024}
}

@article{goh2024b,
  author = {Goh, E. and others},
  title = {Large language model influence on diagnostic reasoning: a randomized clinical trial},
  journal = {JAMA Network Open},
  year = {2024},
  volume = {7},
  pages = {e2440969--e2440969}
}

@inproceedings{wei2022,
  author = {Wei, J. and others},
  title = {Chain-of-thought prompting elicits reasoning in large language models},
  booktitle = {Advances in Neural Information Processing Systems},
  year = {2022},
  volume = {35},
  pages = {24824--24837}
}

@article{chen2025,
  author = {Chen, Q. and others},
  title = {Benchmarking large language models for biomedical natural language processing applications and recommendations},
  journal = {Nature Communications},
  year = {2025},
  volume = {16},
  pages = {3280},
  doi = {10.1038/s41467-025-56989-2}
}

@article{shah2024,
  author = {Shah, S. V.},
  title = {Accuracy, Consistency, and Hallucination of Large Language Models When Analyzing Unstructured Clinical Notes in Electronic Medical Records},
  journal = {JAMA Network Open},
  year = {2024},
  volume = {7},
  pages = {e2425953--e2425953},
  doi = {10.1001/jamanetworkopen.2024.25953}
}

@article{miao2024,
  author = {Miao, J. and others},
  title = {Chain of Thought Utilization in Large Language Models and Application in Nephrology},
  journal = {Medicina},
  year = {2024},
  volume = {60},
  pages = {148}
}

@misc{jaech2024,
  author = {Jaech, A. and others},
  title = {Openai o1 system card},
  howpublished = {arXiv preprint arXiv:2412.16720},
  year = {2024}
}

@misc{guo2025,
  author = {Guo, D. and others},
  title = {Deepseek-r1: Incentivizing reasoning capability in llms via reinforcement learning},
  howpublished = {arXiv preprint arXiv:2501.12948},
  year = {2025}
}

@article{li2024,
  author = {Li, M. and Zhou, H. and Yang, H. and Zhang, R.},
  title = {RT: a Retrieving and Chain-of-Thought framework for few-shot medical named entity recognition},
  journal = {Journal of the American Medical Informatics Association},
  year = {2024},
  volume = {31},
  pages = {1929--1938},
  doi = {10.1093/jamia/ocae095}
}

@misc{zhang2025a,
  author = {Zhang, D. and others},
  title = {MME-CoT: Benchmarking Chain-of-Thought in Large Multimodal Models for Reasoning Quality, Robustness, and Efficiency},
  howpublished = {arXiv preprint arXiv:2502.09621},
  year = {2025},
  url = {https://arxiv.org/abs/2502.09621}
}

@misc{mondillo2025,
  author = {Mondillo, G. and Colosimo, S. and Perrotta, A. and Frattolillo, V. and Masino, M.},
  title = {Comparative evaluation of advanced AI reasoning models in pediatric clinical decision support: ChatGPT O1 vs. DeepSeek-R1},
  howpublished = {medRxiv},
  year = {2025},
  note = {2025.01.27.25321169}
}

@article{wang2025b,
  author = {Wang, Z. and others},
  title = {HealthQ: Unveiling questioning capabilities of LLM chains in healthcare conversations},
  journal = {Smart Health},
  year = {2025},
  volume = {36},
  pages = {100570},
  doi = {10.1016/j.smhl.2025.100570}
}

@article{zhang2019,
  author = {Zhang, Y. and others},
  title = {Does tranexamic acid improve bleeding, transfusion, and hemoglobin level in patients undergoing multilevel spine surgery? A systematic review and meta-analysis},
  journal = {World Neurosurgery},
  year = {2019},
  volume = {127},
  pages = {289--301}
}

@article{yoo2019,
  author = {Yoo, J. S. and Ahn, J. and Karmarkar, S. S. and Lamoutte, E. H. and Singh, K.},
  title = {The use of tranexamic acid in spine surgery},
  journal = {Annals of Translational Medicine},
  year = {2019},
  volume = {7}
}

@article{choi2017,
  author = {Choi, H. Y. and Hyun, S.-J. and Kim, K.-J. and Jahng, T.-A. and Kim, H.-J.},
  title = {Effectiveness and safety of tranexamic acid in spinal deformity surgery},
  journal = {Journal of Korean Neurosurgical Society},
  year = {2017},
  volume = {60},
  pages = {75}
}

@article{cheriyan2015,
  author = {Cheriyan, T. and others},
  title = {Efficacy of tranexamic acid on surgical bleeding in spine surgery: a meta-analysis},
  journal = {The Spine Journal},
  year = {2015},
  volume = {15},
  pages = {752--761}
}

@article{hanada2021,
  author = {Hanada, K. and others},
  title = {Castigating intraoperative bleeding: tranexamic acid, a new ally},
  journal = {Asian Journal of Neurosurgery},
  year = {2021},
  volume = {16},
  pages = {51--55}
}

@article{mallepally2020,
  author = {Mallepally, A. R. and others},
  title = {Use of topical tranexamic acid to reduce blood loss in single-level transforaminal lumbar interbody fusion},
  journal = {Asian Spine Journal},
  year = {2020},
  volume = {14},
  pages = {593}
}

@article{nouri2015,
  author = {Nouri, A. and Tetreault, L. and Singh, A. and Karadimas, S. K. and Fehlings, M. G.},
  title = {Degenerative cervical myelopathy: epidemiology, genetics, and pathogenesis},
  journal = {Spine},
  year = {2015},
  volume = {40},
  pages = {E675--E693}
}

@article{witiw2017,
  author = {Witiw, C. D. and Fehlings, M. G.},
  title = {Degenerative cervical myelopathy},
  journal = {CMAJ},
  year = {2017},
  volume = {189},
  pages = {E116--E116}
}

@article{muhly2016,
  author = {Muhly, W. T. and others},
  title = {Rapid recovery pathway after spinal fusion for idiopathic scoliosis},
  journal = {Pediatrics},
  year = {2016},
  volume = {137}
}

@article{martin2014,
  author = {Martin, C. T. and others},
  title = {Increasing hospital charges for adolescent idiopathic scoliosis in the United States},
  journal = {Spine},
  year = {2014},
  volume = {39},
  pages = {1676--1682}
}

@article{hart1995,
  author = {Hart, L. G. and Deyo, R. A. and Cherkin, D. C.},
  title = {Physician office visits for low back pain: frequency, clinical evaluation, and treatment patterns from a US national survey},
  journal = {Spine},
  year = {1995},
  volume = {20},
  pages = {11--19}
}

@article{gbd2018,
  author = {{Global Burden of Disease Study 2017 Collaborators}},
  title = {Global, regional, and national incidence, prevalence, and years lived with disability for 354 diseases and injuries for 195 countries and territories, 1990-2017: a systematic analysis for the Global Burden of Disease Study 2017},
  journal = {The Lancet},
  year = {2018},
  volume = {392},
  pages = {1789--1858},
  doi = {10.1016/s0140-6736(18)32279-7}
}

@article{hoy2012,
  author = {Hoy, D. and others},
  title = {A systematic review of the global prevalence of low back pain},
  journal = {Arthritis and Rheumatism},
  year = {2012},
  volume = {64},
  pages = {2028--2037},
  doi = {10.1002/art.34347}
}

@misc{wang2025c,
  author = {Wang, W. and others},
  title = {A survey of llm-based agents in medicine: How far are we from baymax?},
  howpublished = {arXiv preprint arXiv:2502.11211},
  year = {2025}
}

@misc{yao2025,
  author = {Yao, Z. and others},
  title = {Are Reasoning Models More Prone to Hallucination?},
  howpublished = {arXiv preprint arXiv:2505.23646},
  year = {2025}
}

@misc{lanham2023,
  author = {Lanham, T. and others},
  title = {Measuring faithfulness in chain-of-thought reasoning},
  howpublished = {arXiv preprint arXiv:2307.13702},
  year = {2023}
}

@misc{shojaee2025,
  author = {Shojaee, P. and others},
  title = {The illusion of thinking: Understanding the strengths and limitations of reasoning models via the lens of problem complexity},
  howpublished = {arXiv preprint arXiv:2506.06941},
  year = {2025}
}

@misc{barez2025,
  author = {Barez, F. and others},
  title = {Chain-of-thought is not explainability},
  howpublished = {Preprint at alphaXiv},
  year = {2025},
  note = {v1}
}

\end{document}